\documentclass{article}

\PassOptionsToPackage{round}{natbib}

 \usepackage[preprint]{preprint}
\usepackage[utf8]{inputenc} 
\usepackage[T1]{fontenc}    
\usepackage[hidelinks]{hyperref}       
\usepackage[font=small, labelfont=bf]{caption} 
\usepackage{url}            
\usepackage{nicefrac}       
\usepackage{microtype}      
\usepackage{enumitem}       
\usepackage{soul}           
\usepackage{booktabs}       
\usepackage{graphicx}
\usepackage{amsmath}
\usepackage{amsfonts}
\usepackage{amssymb}
\usepackage{amsthm}
\usepackage{algorithm} 
\usepackage{algpseudocode}  

\hypersetup{
	colorlinks = true,
	citecolor = blue
}

\newcommand{\ie}{\textit{i}.\textit{e}., }
\newcommand{\eg}{\textit{e}.\textit{g}., }

\newcommand{\aka}{\textit{a.k.a.} }

\newcommand{\norm}[1]{\left\lVert#1\right\rVert}

\newcommand{\diff}{\mathop{}\!\mathrm{d}}

\theoremstyle{definition}

\theoremstyle{plain}

\theoremstyle{remark}

\setuldepth{o}

\newcommand{\NM}{Nelder--Mead}
\newcommand{\iNM}{Iterated Nelder--Mead}

\title{Total Variation Regularization for Compartmental Epidemic Models with Time-Varying Dynamics}

\author{%
  Wenjie Zheng\thanks{www.zhengwenjie.net} \\
  \texttt{contact@zhengwenjie.net} \\
}

\begin{document}

\maketitle

\begin{abstract}
Compartmental epidemic models are among the most popular ones in epidemiology. 
For the parameters (\eg the transmission rate) characterizing these models, 
the majority of researchers simplify them as constants, 
while some others manage to detect their continuous variations. 
In this paper, we aim at capturing, on the other hand, 
\emph{discontinuous} variations, 
which better describe the impact of many noteworthy events, 
such as city lockdowns, the opening of field hospitals, 
and the mutation of the virus, whose effect should be instant. 
To achieve this, we balance the model's likelihood by total variation, 
which regulates the temporal variations of the model parameters. 
To infer these parameters, instead of using Monte Carlo methods, 
we design a novel yet straightforward optimization algorithm, dubbed \iNM, 
which repeatedly applies the \NM{} algorithm. 
Experiments conducted on the simulated data demonstrate that 
our approach can reproduce these discontinuities 
and precisely depict the epidemics.
\end{abstract}

\section{Introduction}
\begin{figure}
	\begin{minipage}{.44\linewidth}
			\centering
			\includegraphics[width=\linewidth]{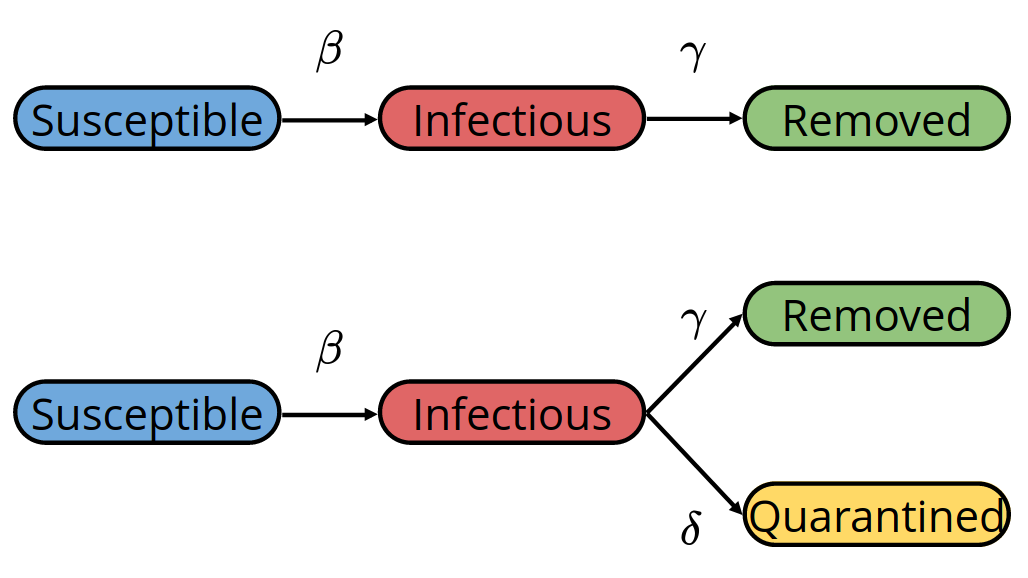}
			\captionof{figure}{Upper: SIR. Lower: SIRQ.}
			\label{fig:model}
	\end{minipage}
	\hfill
	\begin{minipage}{.44\linewidth}
			\centering
			\includegraphics[width=\linewidth]{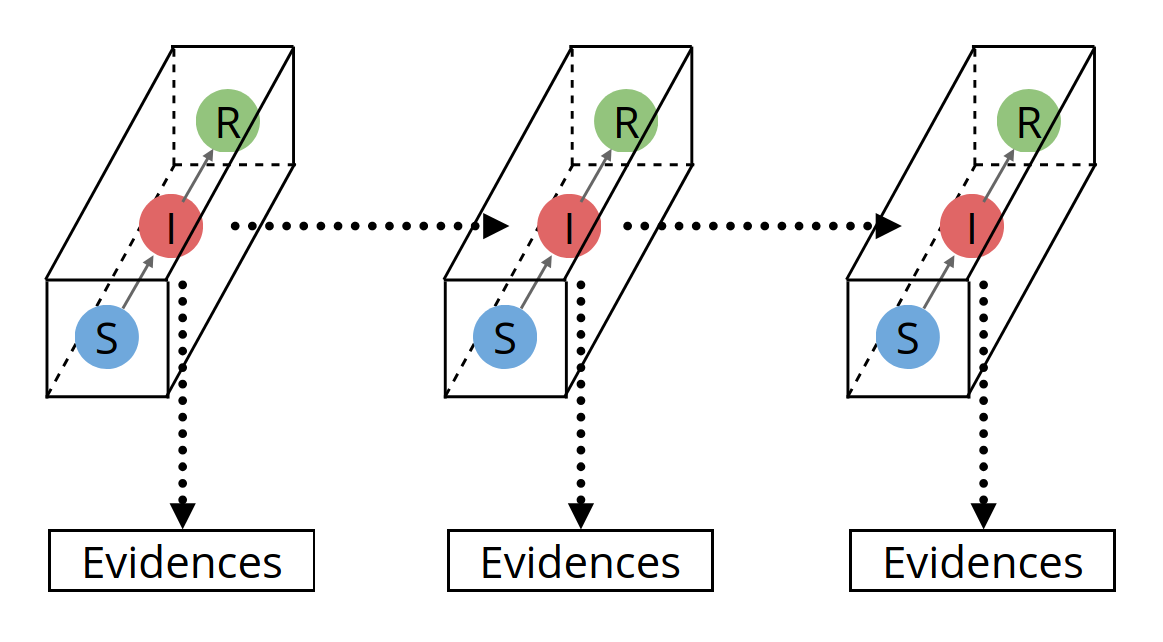}
			\captionof{figure}{A state-space framework feat. SIR.}
			\label{fig:ssm}
	\end{minipage}
\end{figure}

Compartmental models \citep{Kermack1927contribution}, such as SIR, 
are an essential research subject in epidemiology
\citep{Siettos2013mathematical}.
These models divide the population into several \textbf{compartments},
among which each individual transfers according to some \textbf{dynamics}.
For instance, the SIR model prescribes three compartments, 
with the \texttt{Susceptible} compartment for those vulnerable, 
the \texttt{Infectious} compartment for those contagious, 
and the \texttt{Removed} compartment for those 
who either obtained immunity after the recovery or 
died from the infection (Figure~\ref{fig:model}).
The dynamics, on the other hand, govern the mechanics of 
how each individual transfers among these three compartments.

Within this fruitful research branch, a few researchers attempted 
to capture the \emph{time-varying} aspect of the dynamics.
That is, the dynamics are no longer constant ones through 
the entirety of the epidemic; 
rather, they vary due to, among others, changing population behaviors, 
public interventions, seasonal effects, viral evolution.
This approach is not only, obviously, more realistic 
but also more coherent to empirical studies.
For instance, the 1918 influenza pandemic displayed 
``three distinct waves'' of infection within 12 months 
\citep[p.~283]{DaiHa2011mechanistic}.
This kind of phenomenon can be explained only by a time-varying dynamic.

To capture the time-varying aspect, these researchers exclusively supposed 
that the \textbf{parameters} characterizing the dynamics are continuous 
\ul{deterministic functions} or \ul{stochastic processes} of time.
One parameter particularly honored by this privilege is 
the \textbf{transmission rate}, which quantifies 
how often an infectious infects a susceptible.
In the case of continuous deterministic functions, 
the transmission rate has taken the form of exponential functions 
\citep{Chowell2004basic, Althaus2014estimating}, 
sigmoid functions \citep{Camacho2014potential}, 
sinusoid functions \citep{Stocks2018model}, 
cubic B-splines \citep{DaiHa2011mechanistic}, 
and Legendre polynomials \citep{Smirnova2017forecasting}.
In the case of continuous stochastic processes, 
it has taken the form of Wiener process 
\citep{Dureau2013capturing, Funk2018real, Cazelles2018accounting, Kucharski2020early} 
and, more generally, Gaussian processes with the periodic kernel or 
the squared exponential kernel 
\citep{Rasmussen2011inference, Xiaoguang2016bayesian}.

Whilst continuous machinery is useful 
for capturing the time-varying aspect of the dynamics, 
it is not suitable to capture the sudden shocks on the dynamics, 
which entails \emph{discontinuity}.
For instance, during the recent Coronavirus Disease 2019 (COVID-19) pandemic, 
multiple regions (\eg Wuhan, Italy, France) were suddenly closed off.
These measures, especially the draconian one in Wuhan, 
were aimed at instantly reducing the transmission rate.
Modeling them via a continuous function or process 
has the danger of smoothing out the transmission rate shift 
and thus underestimates the efficiency of these measures.

For the accurate detection of such sudden changes in the dynamics, 
we do not impose the property of continuity, 
let alone smoothness on the dynamic.
Instead, the model estimation error is controlled by 
\textbf{total variation regularization}.
Total variation regularization has the effect of detecting discontinuities 
within the investigated object (\eg function, process).
It is used, among others, in image denoising, 
wherein it successfully restores the sharpness of images.
The hence restored object has a well-known \textit{staircase} visual effect.
By applying total variation regularization 
on the calibration of epidemic dynamics, 
we can hope to reconstruct a dynamic mostly constant 
while still allowing some \textit{phase shifts}.

It is worth noting that our approach is fundamentally different from 
the \textit{piecewise} approaches 
(such as the one proposed in \citealp{Funk2017impact}).
The latter artificially break the epidemic into several time periods 
and then models each period individually, 
whilst the former does not presume any locations to implant such breakages 
and, instead, lets the data speak for itself.
The latter are reasonable for the modeling of public interventions, 
which generate \emph{foreseeable} sudden impacts on the epidemic dynamic.
The former is more general and 
can additionally capture \emph{invisible} phase shifts 
induced by, say, viral evolution.
Furthermore, total variation regularization can still be preferable 
for the modeling of foreseeable phase shifts, 
for these shifts may not happen \emph{immediately} after, 
say, the public intervention.
The piecewise approach may neglect the delays 
and thus underestimate the efficiency of the interventions.

To apply total variation regularization in a \emph{principled} way, 
we adopt the well-known \textbf{state-space framework}\footnote{
	Originally named State-Space \ul{Model} (SSM). 
	Here we mimicked several other researchers (such as \citealp{Birrell2018evidence}) and renamed it as a framework to prevent any confusion with the modeling of the dynamics.} 
in epidemiology.
This framework supposes that the underlying compartment status 
is a latent object and hence not directly observable.
To infer the latent status, we can only collect secondary information 
derived from it.
By machine learning terminology, 
it is very similar to the Hidden Markov Model (HMM).
The state-space framework enables the \textbf{evidence synthesis} approach, 
which leverages data from multiple sources: 
Surveillance data (\ie prevalence and incidence) can be supplemented by 
additional serological, demographic, administrative, environmental, 
or phylogenetic data.
Interested readers are referred to \citet[Section~3]{Birrell2018evidence}'s 
review for some examples.
A recent study using phylogenetic data to infer the epidemic of COVID-19 
can be found in \cite{Kucharski2020early}'s work. 

To infer the latent dynamic in the state-space framework, 
researchers almost exclusively adopt some Monte Carlo methods 
such as Sequential Monte Carlo (\aka{particle filter}) or 
particle Markov Chain Monte Carlo (pMCMC).
The procedure starts with the combination of the \ul{prior} 
provided by hypotheses on the latent dynamic and 
the \ul{likelihood} provided by the observation mechanism, 
followed by the simulation of the \ul{posterior} via some Monte Carlo method.
This procedure is so streamlined that an entire software package\footnote{
	\url{https://github.com/StateSpaceModels/ssm}.} 
has been developed for it \citep{Dureau2013ssm}.

We, instead, design an \textbf{\iNM} algorithm 
targeting the \textbf{maximum a posteriori (MAP)} estimate, 
where the objective function of interest is the likelihood 
regularized by total variation.
In contrast to Monte Carlo methods estimating the posterior \ul{mean}, 
MAP focuses on the posterior \ul{mode}.
Past experiences in image denoising suggest that 
MAP restores the discontinuities in the investigated object 
much better than the posterior mean does.
Therefore, the key here is to design a suitable optimization algorithm 
to find the \emph{global} optimum.
Our experiences show that \iNM{} successfully reproduces the discontinuities and precisely depicts the dynamics.

It is worth noting that, due to the nonparametric nature of our approach, 
the objective function under investigation is neither convex nor unimodal. 
This property differentiates our work from many others 
also applying the \NM{} algorithm but on a unimodal objective.
When the objective is unimodal, 
all reasonable descent algorithms all converge to the same optimum 
(ignoring the convergence rate).
Nevertheless, when the objective is multimodal, 
algorithms can be easily trapped in some local optimum,
and it takes much skill to reach the global optimum.
In particular, we discovered that 
the regularization hyperparameter controls not only the bias-variance tradeoff 
of the model but also the topology of the objective function.
That is, an overly small value or an overly large value 
can both render the objective function challenging to optimize 
and hence trap the optimization algorithm. 

The contributions of this study can be divided into two groups.
\begin{itemize}
\item It is the first one to propose total variation regularization 
to capture the discontinuity of time-varying epidemic dynamics. 
Moreover, it is the first one to apply the nonparametric approach 
simultaneously on more than one parameter.
In the study, we focus on SIR and an \textit{ad hoc} variant SIRQ, 
but, thanks to the state-space framework adopted, 
our methodology can be extended to other compartmental models 
and even non-compartmental ones.

\item It is the first one successfully using \emph{non}-Monte Carlo methods 
for the inference of nonparametric compartmental models.
The \iNM{} algorithm designed for this purpose 
reveals the role of hyperparameters 
in tuning the topology of the objective function.
Such knowledge might help solve the mysteries 
in the training of deep neural networks.

\end{itemize}

This paper is organized as follows.
Section~\ref{sec:compartment} introduces the most classic compartmental model, 
\textbf{SIR}, as well as designs a variant, \textbf{SIRQ}.
The former describes an environment without interventions, 
whilst the latter adds an additional compartment, \texttt{Quarantined}, 
to model the intervention.
Section~\ref{sec:ssm} describes the state-space framework, 
whose two components are the state equation and the observation equation.
Section~\ref{sec:tv} presents total variation regularization and \iNM.
Section~\ref{sec:experiment} tests our approach on simulated data, 
and the results are promising.

\section{Compartmental models: SIR and SIRQ}\label{sec:compartment}

SIR, standing for \texttt{Susceptible-Infectious-Removed}, 
is the basic compartmental model.
By using SIR as a stepping stone, 
we can understand more complex models such as SIRQ, 
which is promoted as a novel model in this paper.

\subsection{SIR model}\label{sub:sir}
The SIR model separates the population into three compartments: 
\texttt{susceptible}, \texttt{infectious}, and \texttt{removed}.
Each individual (logically) transfers among these compartments 
according to his health status (Figure~\ref{fig:model}).
\begin{description}
\item[Susceptible:] The population in this compartment 
are healthy people who are vulnerable to the disease.
\item[Infectious:] The population in this compartment 
are infected people who are free to infect those susceptible.
\item[Removed:] This compartment consists of two groups of people -- those who 
died from the disease or recovered from it and hence obtained immunity.
\end{description}

To describe the dynamic governing the mechanism, 
two types of setups are possible -- stochastic or deterministic. 
The stochastic one assumes that 
each individual transfers according to some probability.
The usual setup is that the probability of a healthy people 
transferring from \texttt{susceptible} to \texttt{infectious} 
follows an exponential distribution of parameter $\beta$, 
and the probability of an infected people 
transferring from \texttt{infectious} to \texttt{removed} 
follows an (independent) exponential distribution of parameter $\gamma$.

On the other hand, the deterministic one simplifies the above process 
by taking advantage of the law of large numbers.
Instead of studying the dynamic on an individual basis, 
the deterministic setup considers it at the aggregate level.
That is, the number of individuals in each compartment 
varies according to some ordinary differential equations (ODE).

Let $S_t$, $I_t$, and $R_t$ be the number of 
susceptible, infectious, and removed at time~$t$, respectively.
Let $N_t = S_t + I_t + R_t$ be the number of the total population, 
which is supposed to be constant 
(\ie there is no newborn or death owing to reasons 
other than the infectious disease in question).
Then the stochastic dynamic can be expressed by the following equations.
\begin{align*}
& \Pr(S_{t+h} - S_t = -1, I_{t+h} - I_t =  1 | S_t, I_t, R_t) = \beta S_t I_t h / N_t + o(h), \\
& \Pr(I_{t+h} - I_t = -1, R_{t+h} - R_t =  1 | S_t, I_t, R_t) = \gamma I_t h + o(h).
\end{align*}
The deterministic dynamic can be expressed by the following ODE.
\begin{align*}
& \frac{\diff S_t}{\diff t} = - \frac{\beta S_t I_t}{N_t}, \\
& \frac{\diff I_t}{\diff t} =   \frac{\beta S_t I_t}{N_t} - \gamma I_t, \\
& \frac{\diff R_t}{\diff t} =   \gamma I_t.
\end{align*}
For large-scale epidemics, the deterministic dynamic is 
a good enough approximation of the stochastic dynamic 
\citep{Kurtz1987approximation}.
Interested readers are also referred to 
\citet[p.~301]{Siettos2013mathematical} for a quick review 
and to \citet[Section~3.2]{Birrell2018evidence} for a detailed discussion.
%

The parameter $\beta$ is called the transmission rate, 
and $\gamma$ is called the removal rate.
The ratio $\beta / \gamma$ is associated with 
the most crucial quantity of infectious diseases, 
the \textbf{basic reproduction number} $\mathcal{R}_0=\frac{\beta}{\gamma}$, 
which stands for the average number of victims 
an infectious is expected to infect at the very beginning of the outbreak.
If $\mathcal{R}_0>1$, the disease becomes an epidemic; 
if $\mathcal{R}_0<1$, the disease dies out; 
if $\mathcal{R}_0=1$, it is an endemic 
(\ie the number of infectious neither grows nor deceases).

\subsection{SIRQ model}\label{sub:sirq}
There are many extensions to SIR.
Here, we design another one, 
dubbed \texttt{susceptible-infectious-removed-quarantined} (SIRQ), 
to include the influence of public interventions.
SIRQ creates a fourth compartment, \texttt{quarantined}, 
which hosts the part of infectious getting quarantined or hospitalized 
(Figure~\ref{fig:model}).
Therefore, the infectious can have two futures: 
either they stay wild and get nature-selected (\ie removed) 
as in the SIR model, or they get quarantined, 
which prevents them from infecting others all the same.

The dynamics of SIRQ are very similar to the above,
so we present here only the deterministic one.
\begin{align*}
& \frac{\diff S_t}{\diff t} = - \frac{\beta S_t I_t}{N_t}, \\
& \frac{\diff I_t}{\diff t} =   \frac{\beta S_t I_t}{N_t} - \gamma I_t - \delta I_t, \\
& \frac{\diff R_t}{\diff t} =   \gamma I_t, \\
& \frac{\diff Q_t}{\diff t} =   \delta I_t,
\end{align*}
where $Q_t$ is the number of quarantined at time $t$, 
$N_t = S_t + I_t + R_t + Q_t$ is the total number of the population, 
and $\delta$ is the rate the infectious getting quarantined.
The ratio $\gamma / \delta$ reflects 
the ratio of the infectious staying under the radar.
Here, the quarantined are assumed not infectious 
(we consider the possibility of infecting healthcare personnel 
to be neglectable).
Also, we do not further specify 
the outflow of the \texttt{quarantined} compartment, 
so it contains three types of people -- 
those being quarantined, those who died during the quarantine, 
and those who recovered during the quarantine.

This model is particularly relevant to the situation of COVID-19.
On the one hand, many victims of COVID-19 are asymptomatic.
They will hence not go to the hospital, 
and they get recovered all by themselves.
On the other hand, given the high transmissibility of COVID-19, 
there are not enough hospital resources for every patient.
Many infectious have to stay wild and get nature-selected.
Besides the above two reasons, there is another one specific to China, mainland:
the \ul{recall} of the test kits is unsatisfying.

One usage of this model is to speculate the ratio of asymptomatic patients 
or the ratio of hospitalization.
The \texttt{removed} compartment is supposed to be unobservable, 
whilst the \texttt{quarantined} compartment can be observed 
from the \ul{confirmed cases}.
Our experiments show that, in the parametric setting, 
the ratio of asymptomatic patients or the like can be accurately inferred 
given only sparse information on the infectious and the quarantined.

Concerning the basic reproduction number, it has two choices.
The controlled version uses $\frac{\beta}{\gamma+\theta}$, 
whose value determines whether the disease will become an epidemic or die out.
The uncontrolled version uses $\frac{\beta}{\gamma}$, 
which stands for the outcome if the quarantine measure is ever called off.

\section{State-space framework}\label{sec:ssm}
The state-space framework has become the \textit{de facto} state of the art 
for the usage of compartment models.
Many studies, such as the one by \citet{Wu2020nowcasting}, 
use this framework implicitly.
This section will first lay down 
a solid mathematical foundation of the state-space framework
and then complement it with some concrete examples. 

\subsection{Mathematical foundation}\label{sub:foundation}

The state-space framework is very similar 
to the Hidden Markov Model in machine learning.
It assumes that the compartment status is in the invisible state space 
and that we can only observe secondary information derived from the states.
Let us illustrate this concept with SIR.
The vector $(S_t, I_t, R_t)$ forms the (invisible) state at time $t$.
The dynamic characterized by the parameters $(\beta, \gamma)$ 
connects the temporally successive states 
(\eg $(S_t, I_t, R_t)$ and $(S_{t+1}, I_{t+1}, R_{t+1})$).
The state at time $t$ determines the evidence data 
that we could collect at time $t$ (Figure~\ref{fig:ssm}).

In the state-space framework, there are usually two tasks.
The first one is to infer the underlying state status, 
which tells us the severity of the current epidemic.
The second one is to characterize the properties (\eg $\mathcal{R}_0$) 
of the epidemic itself, 
which helps us evaluate the ferocity of our enemy.
These two tasks are interdependent: the fulfillment of one entails the other.

\begin{figure}
	\centering
	\includegraphics[width=0.3\linewidth]{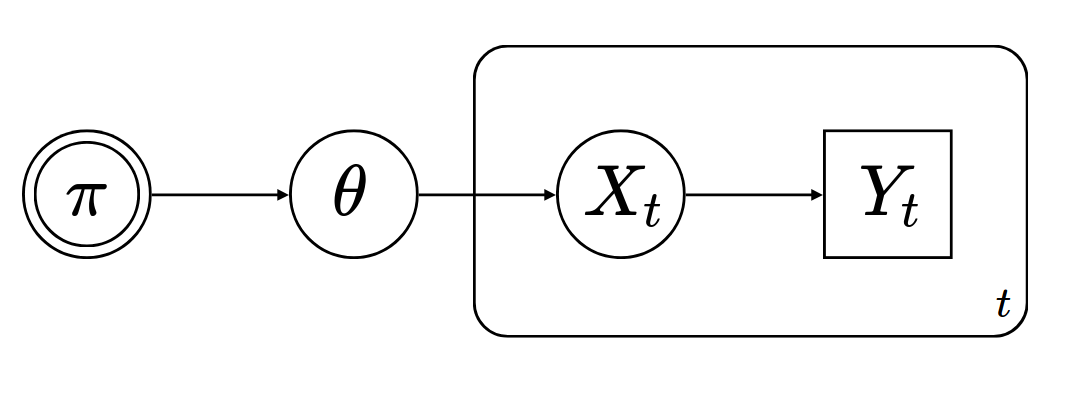}
	\caption{Graphical model for the state-space framework.}
	\label{fig:dag}
\end{figure}

It is a principled way to infer these two tasks together 
through the \ul{graphical model} (Figure~\ref{fig:dag}).
First of all, 
the epidemic dynamic parameter $\theta$ (in SIR, $\theta=(\beta, \gamma)$) 
governs the temporal transition of the states $X_t$ 
(in SIR, $X_t=(S_t, I_t, R_t)$):
\[
X_{t+1} | X_t \sim p_\theta(\cdot | X_t) \quad \text{(state equation)}.
\]
This probability distribution can be degenerate, 
in which case it is reduced to a \emph{deterministic} epidemic dynamic.
SIR and SIRQ provide examples of $p_\theta$.
Next, 
the invisible states $X_t$ generates some empirical evidence, denoted as $Y_t$:
\[
Y_t | X_t \sim p_{\bar\eta} (\cdot | X_t)  \quad \text{(observation equation)},
\]
where $\bar\eta$ is often manually selected by the researchers 
to prevent any unintended complexity.
Also, Bayesian statisticians may go further by supposing that 
the parameter $\theta$ lies in some probability space.
Thus, they will include in the model a prior:
\[
\theta \sim \bar\pi(\cdot)  \quad \text{(prior)},
\]
where the prior $\bar\pi$ is preset.
Finally, to infer the state $X_t$ and the parameter $\theta$, 
we can build their posterior distribution, 
which is proportional to the joint distribution.
Let $x_i$ be the value observed of $X_i$ 
and let $x_{1:T}$ denote the tuple $x_1, \ldots, x_T$, 
then the posterior distribution
\[
p(\theta, x_{1:T} | y_{1:T}) \propto p_{\bar\eta}(y_{1:T}|x_{1:T}) p_\theta(x_{1:T}) \bar\pi(\theta)  \quad \text{(posterior)}.
\]

The above assumes the Markov property.
In practice, the state-space framework can be more general 
by abandoning the Markov property or by introducing dependence between $Y_t$.


In the following, we will give some concrete examples.
Section~\ref{sub:observation} gives some candidates 
for the observation equation,
and Section~\ref{sub:prior} gives some candidates for the prior.
As for the state equation, 
SIR and SIRQ in the previous section are good candidates.

\subsection{Candidates for the observation equation}\label{sub:observation}
The observation equation introduces evidence 
for the inference of the dynamic and the latent states.
Traditionally we use only surveillance data, 
but now an increasing number of studies start to use exotic data 
such as serological, demographic, administrative, environmental, 
and phylogenetic data \citep{Birrell2018evidence}.

The first example is how researchers used traveling data 
to infer the total number of COVID-19 patients in Wuhan, China.
In January 2020, there are a large number of people 
infected by COVID-19 in Wuhan city.
Unaware of the infection, they traveled abroad and later got diagnosed.
\citet{Imai2020estimating} modeled this natural experiment 
as sampling with replacement.
That is, the number of patients diagnosed abroad 
follows a binomial distribution $\text{Bin}(m, I/N)$, 
where $m$ stands for the total number of outbound travelers, 
$I$ the number of infectious, and $N$ the total population.
Rigorously speaking, the natural experiment is more like 
sampling \emph{without} replacement, 
but the difference is neglectable when $m \ll N$.
\citet{Wu2020nowcasting} modeled it, alternatively, 
as a Poisson distribution $\text{Poi}(mI/N)$.
Incidentally, there is no fundamental difference between the above two options, 
for $\text{Bin}(m, I/N) \approx \text{Poi}(mI/N)$ 
when $m>20$ and $I \ll N$ thanks to the \ul{law of rare events}.

The second example is based on a probably unpopular opinion:
using the number of confirmed cases 
to infer the \texttt{quarantined} compartment.
It is attempting to use the confirmed cases 
for the \texttt{infectious} compartment instead.
Nonetheless, a smarter and more sensible arrangement is 
to use them for the \texttt{removed} compartment in SIR 
or the \texttt{quarantined} compartment in SIRQ.
When a person is confirmed infected, 
he most likely will be admitted to the hospital 
and thus lose the ability to infect others 
(if the risk of infecting the healthcare personnel is low).
The \texttt{removed} compartment in SIR is used for this purpose.
If the hospitalization is imperfect 
(\ie a part of patients are not confirmed and have to be nature-selected), 
this is a perfect scenario for the SIRQ model, 
where the confirmed cases can be associated with 
the \texttt{quarantined} compartment.
\citet[Section~5.2]{Xiaoguang2016bayesian} thought alike 
and used the confirmed cases to infer the \texttt{removed} compartment.
In this article, we will use the confirmed cases 
for the \texttt{quarantined} compartment.
The observation equation can 
either feature the Gaussian distribution or the Poisson distribution.
In fact, there is no significant difference, 
as $\texttt{Poi}(\lambda) \approx \mathcal{N}(\lambda, \lambda)$ 
for sufficiently large $\lambda$.

The third example is to use the serological data to infer 
the \texttt{removed} compartment by detecting the antibody in the blood.
In an imperfect quarantine scenario modeled by SIRQ, 
some patients survived the disease without formal medical intervention 
or died because of it.
These patients are never administratively confirmed, 
but they reflect the severity of the epidemic all the same. 
To fairly evaluate the epidemic, 
it is essential to estimate the portion of unconfirmed cases.
Since recovered people will have an antibody in the blood, 
the serological data can help us screen out this group of people.
To get an unbiased estimate, we can sample the whole population, 
and then it is reduced to an elementary statistics problem.

\subsection{Candidates for the prior}\label{sub:prior}
The prior denotes the preset distribution on 
the parameter characterizing the dynamic 
(the parameter characterizing the observation equation is preset).
This parameter can be finite-dimensional (vector) 
or infinite-dimensional (function).
The finite-dimensional cases usually use an uninformative prior (\ie constant), 
and the posterior degenerates to the plain likelihood.
It is only in the infinite-dimensional cases 
that the selection of prior becomes nontrivial.

In the infinite-dimensional case, the parameter $\theta_t$ is time-varying.
We are to sample functions for $\theta_t$ in some function space.
In other words, $\theta_t$ is a stochastic process.
There are mainly two candidate spaces for this purpose.
The first defines the stochastic process by a stochastic differential equation:
\[
\diff h(\theta_t) = \mu_{t, \theta} \diff t + \sigma_{t, \theta} \diff B_t,
\]
where $\mu_{t, \theta}$ is the drift, 
$\sigma_{t, \theta}$ is the volatility, 
$B_t$ is a standard Wiener process, 
and $h(\cdot)$ is a preset deterministic function.
The most common choice for $\theta_t$ is Brownian motion where $h(\cdot)=\cdot$ 
and geometric Brownian motion where $h(\cdot)=\log(\cdot)$.
If the process is expected to converge, 
\citet[p.~4]{Dureau2013capturing} also proposed the Ornstein Uhlenbeck process.

The second function space is the Gaussian process.
A Gaussian process has the property that 
its arbitrary segments follow a multivariate Gaussian distribution, 
hence the name.
It is wildly used in nonparametric Bayesian statistics.
The distribution of the Gaussian process is uniquely defined 
by its expectation (function) and 
its kernel (covariance function) $K(\cdot, \cdot)$.
\citet{Xiaoguang2016bayesian} investigated two types of kernels: 
squared exponential 
\[
K(x, y) = \alpha^2 \exp[-(x-y)^2/(2\ell^2)]
\]
and periodic
\[
K(x, y) = \alpha^2 \exp(-\ell^{-1}(1-\cos(2\pi\omega^{-1}|x-y|))).
\]

In the above description, the parameter $\theta_t$ is quite abstract.
In practice, this $\theta_t$ has concrete meanings.
For example, in SIRQ, $\theta_t=(\beta_t, \gamma_t, \delta_t)$, in which case, 
we can apply independent priors individually on each component.

\section{Total variation regularization and \iNM}\label{sec:tv}
Priors can limit the model complexity and hence 
control the model estimation error in the context of bias-variance tradeoff.
An alternative approach is to apply a regularization on the log-likelihood.
This section introduces the concept of \textbf{total variation regularization}, 
which is popular in image denoising.
It has the advantage of 
detecting the discontinuities in the investigated object, 
and thus it is expected here to capture sudden shocks on the dynamics.
Also, in contrast to Monte Carlo methods, 
the standard approach to calculate the posterior mean,
this section designs a novel algorithm, dubbed \textbf{\iNM}, 
intended to calculate the posterior mode.

\subsection{Total variation regularization}\label{sub:tvr}
Section~\ref{sec:ssm} formulates the posterior as 
\[
p_{\bar\eta}(y_{1:T}|x_{1:T}) p_\theta(x_{1:T}) \bar\pi(\theta). 
\]
By applying logarithm, we get
\[
\underbrace{\log p_{\bar\eta}(y_{1:T}|x_{1:T}) + \log p_\theta(x_{1:T})}_{\text{log-likelihood}} +
\underbrace{\log \bar\pi(\theta)}_{\text{regularization}},
\]
where the last term can be regarded as a regularization.
In other words, prior is \emph{one} type of regularization.

An alternative regularization would be substituting the prior for a norm applied on $\theta$:
\[
\underbrace{\log p_{\bar\eta}(y_{1:T}|x_{1:T}) + \log p_\theta(x_{1:T})}_{\text{log-likelihood}} +
\underbrace{\norm{\theta}}_{\text{regularization}}.
\]
This norm can be, among others, total variation
\[
\norm{\theta}_\text{TV} := \sup \sum_i |\theta_{t_{i+1}} - \theta_{t_i}|,
\]
where the supreme runs over the set of all \ul{partitions},
or \textbf{quadratic variation}
\[
[\theta] := \lim_{\norm{P}\rightarrow0} \sum_i (\theta_{t_{i+1}} - \theta_{t_i})^2,
\]
where $P$ ranges over all partitions and the norm is the \ul{mesh}.

The above frames the prior as a type of regularization; 
the converse is true as well.
Total variation regularization or quadratic variation regularization 
can be regarded as a prior on the space of functions 
with finite total variations or finite quadratic variations, respectively.
In particular, quadratic variation regularization 
is equivalent to specifying $\theta$ as Brownian motion.

\subsection{\iNM}\label{sub:inm}
It is wildly perceived, in the computer vision community, 
that the posterior mean's ability to recover the discontinuity 
is not as good as the posterior mode.
This subsection describes the challenges to calculate the posterior mode 
as well as provides solutions to it.

The first challenge is the unavailability of the gradient.
Although the part of the likelihood associated with the observation equation 
can be easily differentiated, 
the part associated with the state equation does not have closed-forms.
Therefore, we have to rely on some zero-order algorithm.
One algorithm fit for this purpose is the \textbf{\NM} method 
(\aka \textbf{downhill simplex method}), 
which is prevalent in civil engineering.
For example, to build a suspension bridge, 
an engineer has to choose the thickness of each strut, cable, and pier.
These elements are interdependent, 
and it is not easy to determine the impact of changing any specific element.
Analogously, in our context, 
changing the value of the dynamic at any single point 
is unlikely to have a significant impact on the whole epidemic 
(the integral does not depend on the function values on a \ul{null set}).

The second challenge is the \textbf{multimodality} of the objective function, 
which refers to functions with multiple modes.
Two factors contribute to this multimodality.
On the one hand, when the observation is sparse, 
there are many candidate dynamics able to reproduce the observation accurately; 
each candidate then forms a valley.
On the other hand, 
constant dynamics do not suffer from the regularization penalty, 
and the modification on a single point will not affect the overall dynamic 
(null set) but does increase total variation, 
so each constant dynamic also forms a valley.
The optimization algorithm can, therefore, 
be easily trapped in some local optimum.
To mitigate the influence of multimodality, 
we designed the \textbf{\iNM} algorithm, 
which repeatedly reruns \NM{} on the new local optimum. 
(Algorithm~\ref{alg:inm}).

\begin{algorithm}
	\caption{\iNM}
	\label{alg:inm}
	\begin{algorithmic}
	\Require initiate point $\textbf{x}$, objective function $\textbf{f}$
	\Repeat
		\State $\textbf{x} \gets \text{\NM}(\textbf{f}, \textbf{x})$
	\Until{\text{stop condition fulfilled}}
	\end{algorithmic}
\end{algorithm}

\begin{figure}
	\centering
	\includegraphics[width=\linewidth]{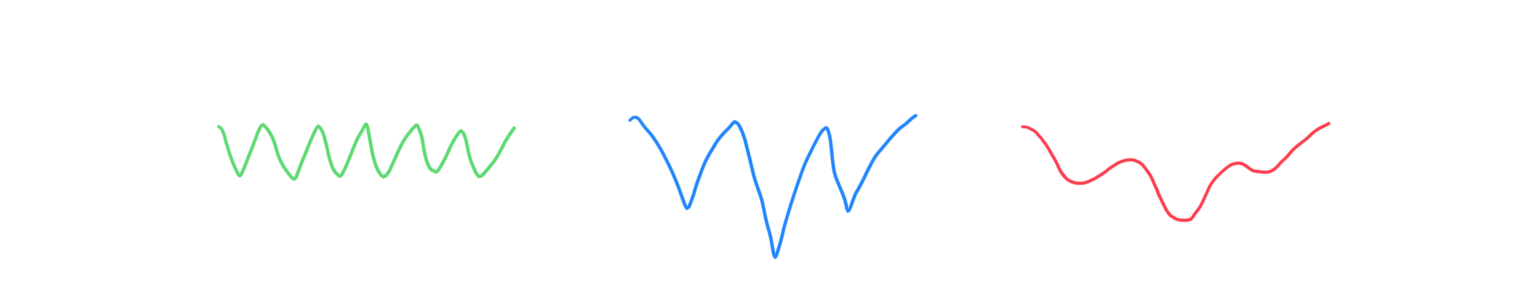}
	\caption{Topology of the objective function with various regularization weights. Left: under-regularized. Middle: over-regularized. Right: well-regularized.}
	\label{fig:topology}
\end{figure}

\iNM{} mitigates the problem of multimodality to some extent, 
but it does not affect the hardship of the problem itself.
During the experiments, we discovered that 
the regularization hyperparameter (weight) plays a critical role 
in defining the hardship of the problem 
by altering the topology of the objective function.
Indeed, when the regularization weight is small, 
the objective function has many equally good local minima---the 
desired local minimum hides among its peers. 
When the regularization weight is large, 
the objective function contains fewer local minima 
with each minimum, however, being an abyss. 
Once the solver loses its way into one abyss, 
it has no chance ever to escape. 
Therefore, the regularization weight should be something 
in-between---highlight the desired local minimum 
while preserving the smoothness of the objective function 
(Figure~\ref{fig:topology}).

\section{Experiments}\label{sec:experiment}
Three types of dynamics and three types of (simulated) data 
are investigated in this study.
The three types of dynamics are 
constant SIRQ with no regularization, 
time-varying SIR with regularization, and 
time-varying SIRQ with regularization.
All dynamics use the ODE versions 
specified in Section~\ref{sec:compartment} 
and are implemented with the \ul{Euler-Maruyama scheme}.

The three types of data are virulence data, surveillance data, 
and serology data:
\begin{description}
	\item[Virulence.] Sample the population 
	and test for the pathogen (\eg virus). 
	This data is for the estimation of 
	the current number of \texttt{infectious} people
	\item[Surveillance.] The confirmed and thus quarantined cases 
	of the disease. 
	This data is for the estimation of 
	the number of \texttt{removed} (in SIR) or 
	\texttt{quarantined} (in SIRQ) people.
	\item[Serology.] Sample the population and test for the antibody. 
	This data is for the estimation of the \texttt{removed} people in SIRQ 
	provided that the infectious disease in question is a novel one.\footnote{
		Otherwise, this data should be used 
		not only for the \texttt{removed} 
		but also for the \texttt{immune} compartment 
		in an \ul{MSIR} model.}
	If the sampling cost is a concern, 
	this data can be collected along with the virulence data.
\end{description}


\subsection{Constant SIRQ}
The SIRQ model has three parameters $\theta = (\beta, \gamma, \delta)$.
The epidemic takes place in a small town with 1000 inhabitants.
It starts with 10 \texttt{infectious} people 
and 0 \texttt{removed} or \texttt{quarantined} people.
Then the epidemic develops with a dynamic 
$(\beta, \gamma, \delta) = (0.3, 0.03, 0.07)$.
Two types of evidence are available for the inference of the epidemic.
The virulence data is 9 samples during the life of the epidemic, 
with a sample size of 10 people each.
The surveillance data is the number of confirmed cases at 8 different moments.
The \ul{maximum likelihood estimate} yields the value 
$(\hat\beta, \hat\gamma, \hat\delta) = (0.307, 0.030, 0.073)$, 
very close to the ground truth.

Although this experiment does not include any time-varying factor, 
it demonstrates the power of this simple model. 
With virulence data and surveillance data only, 
we can precisely estimate the basic reproduction number 
and the quarantine ratio $\delta/\gamma$.
This model is particularly useful in the COVID-19 pandemic, 
where many victims are asymptomatic.

\begin{figure}
	\begin{minipage}{.48\linewidth}
		\centering
		\includegraphics[width=.48\linewidth]{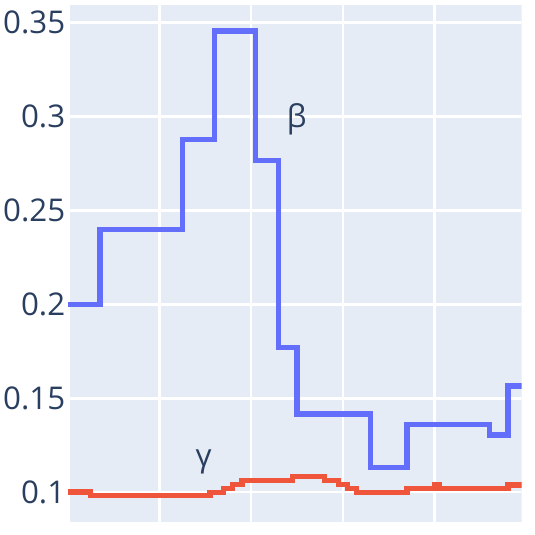}
		\includegraphics[width=.48\linewidth]{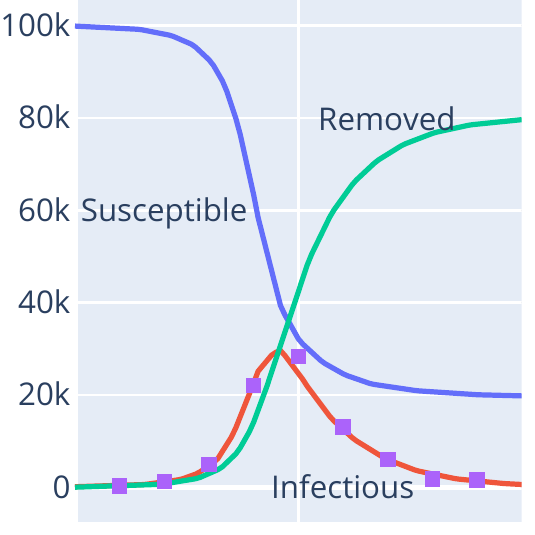}
		\includegraphics[width=.48\linewidth]{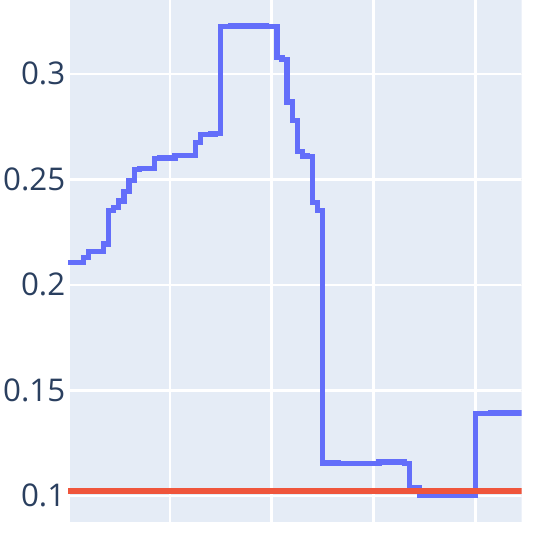}
		\includegraphics[width=.48\linewidth]{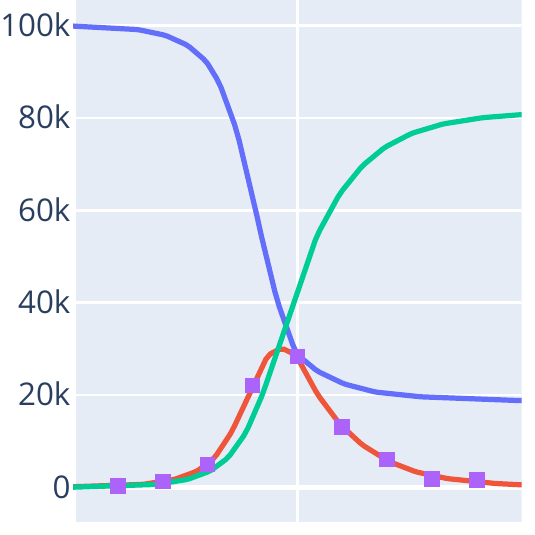}
		\captionof{figure}{Time-varying SIR. Left: the dynamic. 
			Right: the resulted epidemic and the evidence 
			(represented by the dots). 
			Top: ground truth. Bottom: estimation. 
			$\beta$ denotes the transmission rate; 
			$\gamma$ denotes the removal rate.}
		\label{fig:sirdynamic}
	\end{minipage}
	\hfill
	\begin{minipage}{.48\linewidth}
		\centering
		\includegraphics[width=.48\linewidth]{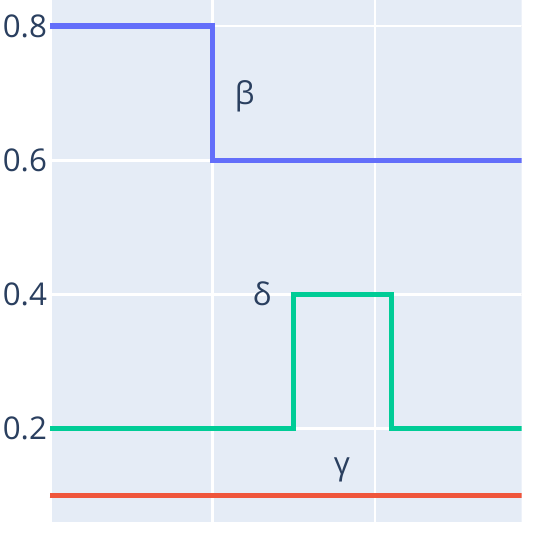}
		\includegraphics[width=.48\linewidth]{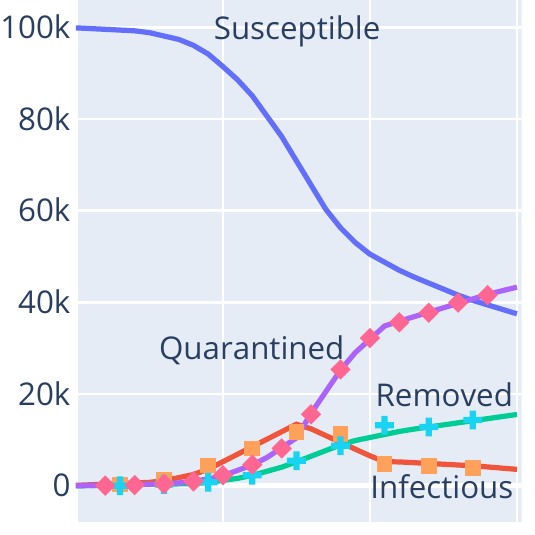}
		\includegraphics[width=.48\linewidth]{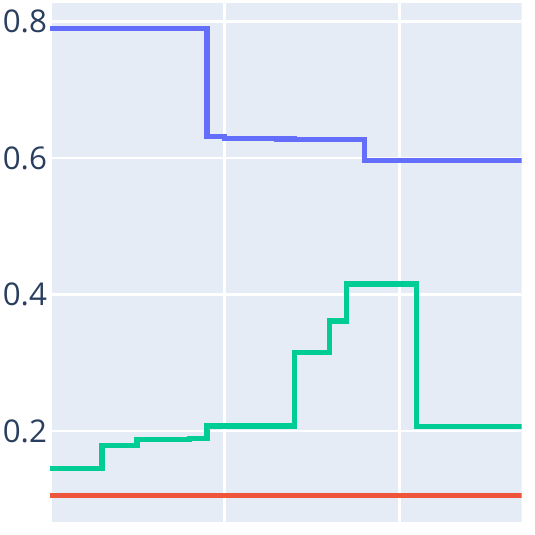}
		\includegraphics[width=.48\linewidth]{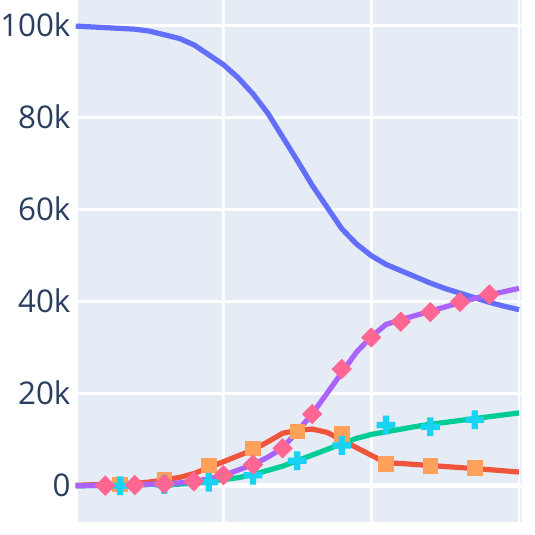}
		\captionof{figure}{Time-varying SIRQ. Left: the dynamic. 
			Right: the resulted epidemic and the evidence (dots). 
			Top: ground truth. Bottom: estimation. 
			$\beta$ denotes the transmission rate; 
			$\gamma$ denotes the removal rate; 
			$\delta$ denotes the quarantine rate.}
		\label{fig:sirqdynamic}
	\end{minipage}
\end{figure}

\subsection{Time-varying SIR with regularization}
The model tested in this subsection has one time-varying parameter 
and one constant parameter $\theta_t = (\beta_t, \gamma)$.
The transmission rate $\beta_t$ is supposed to be time-varying 
to reflect, among others, the  gatherings in holidays, 
the adoption of social distancing.
The removal rate $\gamma$ is supposed to be constant 
because the aggressiveness of the virus and the resistance of the population 
are believed to be stable (unless the virus mutates).
In the experiment, the ground truth $\beta_t$ firstly increases 
because of holiday gatherings.
It then drops because of increasing public awareness
(Figure~\ref{fig:sirdynamic}).

The epidemic takes place in a city of 100k inhabitants.
It starts with 100 \texttt{infectious} people and 0 \texttt{removed} people. 
The only data available to infer the epidemic is the 9-sample virulence data 
during the life of the epidemic, with each sample containing 1k people.
The lack of the number of confirmed cases suggests that 
this is a foreign country trying to evaluate an epidemic-struck country 
with information censorship; 
the virulence data corresponds to the exported cases.

The data is considered \emph{sparse} 
in contrast to the 100-dimensional parameter, 
which justifies the necessity of regularization.
We applied total variation regularization and solved it with \iNM.
The estimated dynamic qualitatively recovers the ground truth and
generates evidence identical to the sample collected 
(Figure~\ref{fig:sirdynamic}).
 
\subsection{Time-varying SIRQ with regularization}
The model tested in this subsection has 
\emph{two} time-varying parameters and one constant parameter 
$\theta_t = (\beta_t, \gamma, \delta_t)$.
The transmission rate $\beta_t$ is supposed to be time-varying 
to reflect, among others, the  gatherings in holidays, 
the adoption of social distancing.
The removal rate $\gamma$ is supposed to be constant 
because the aggressiveness of the virus and the resistance of the population 
are believed to be stable (unless the virus mutates).
The quarantine rate $\delta_t$ is supposed to be time-varying 
to reflect the opening of field hospitals.
In the experiment, the ground truth $\beta_t$ is mostly constant
with one sudden drop because of the city lockdown, 
whilst the ground truth $\delta_t$ rises suddenly 
thanks to the opening of a field hospital 
and then drops to the previous level because the hospital is full 
(Figure~\ref{fig:sirqdynamic}).

The epidemic takes place in a city of 100k inhabitants.
It starts with 100 \texttt{infectious} people 
and 0 \texttt{removed} or \texttt{quarantined} people. 
The data available here is much richer than the scenario in the last subsection.
It includes a 9-sample virulence data with each containing 1k people, 
the confirmed cases, 
and a 9-sample serology data with each also containing 1k people.
All this information is not too difficult to obtain 
for countries with excellent governance.

One challenge to infer this model is 
the coexistence of \emph{two} time-varying parameters.
Still, total variation regularization and \iNM{} 
succeeded in reproducing the ground truth, 
and the estimated dynamic generates evidence that perfectly fits the reality.

\section{Conclusions}\label{sec:conclusion}
The combination of total variation regularization and \iNM{} successfully detects the discontinuities of the underlying time-varying dynamics.
When the epidemic follows an SIR dynamic 
with time-varying transmission rate $\beta_t$, 
the proposed combination recovers the dynamic 
with the help of sparse virulence data.
When the epidemic follows an SIRQ dynamic 
with time-varying transmission rate $\beta_t$ 
and time-varying quarantine rate $\delta_t$, 
the proposed combination recovers the dynamic 
with the help of virulence, surveillance, and serology data.

There are three directions to strengthen this study.
Firstly, 
it is unclear whether the local optimum achieved by \iNM is the global one.
Since the optimization algorithm plays a crucial role 
in the solution-finding process, 
one part of the research effort should be 
improving the searching ability of the optimization algorithm.
Secondly, it is currently unclear 
whether these dynamics could be \emph{perfectly} recovered 
with more data or under better conditions.
One research direction thus consists of 
precisely inferring the underlying dynamics 
so that public policies (\eg city lockdown) could be better evaluated.
It also helps compare the efficiency of different policies in the same country 
and the efficiency of the same policy in different countries.
Thirdly and most importantly, 
we believe that our machinery is a handy weapon against COVID-19.

%

\small
\bibliographystyle{agsm}
\bibliography{epidemic}

\end{document}